\documentclass[11pt]{article}

\usepackage[final]{acl}

\usepackage{times}
\usepackage{latexsym}

\usepackage[T1]{fontenc}
\usepackage[utf8]{inputenc}

\usepackage{microtype}

\usepackage{inconsolata}

\usepackage{graphicx}
\usepackage{comment}
\usepackage{array, siunitx}
 \usepackage{float}

\usepackage{multirow}
\usepackage{booktabs} % For thicker horizontal lines
\newcolumntype{P}[1]{>{\centering\arraybackslash}p{#1}}
\usepackage{makecell}
\usepackage{colortbl}
\usepackage{tabularx}
\usepackage{longtable}
\usepackage{amsmath} % for the \eqref command
\title{ELLIS Alicante at CQs-Gen 2025: Winning the critical thinking questions shared task: LLM-based question generation and selection}

\author{
    \textbf{Lucile Favero\textsuperscript{1}},
    \textbf{Daniel Frases \textsuperscript{2}} \thanks{Worked performed during an internship at ELLIS Alicante.},
    \textbf{Juan Antonio Pérez-Ortiz\textsuperscript{3}},\\
    \textbf{Tanja Käser\textsuperscript{4}}, 
   \textbf{ Nuria Oliver\textsuperscript{1}}
\\
    \textsuperscript{1} ELLIS Alicante, Spain,
    \textsuperscript{2} Universidad Alfonso X el Sabio, Spain,\\
    \textsuperscript{3} Universitat d'Alacant, Spain,
    \textsuperscript{4} École Polytechnique Fédérale de Lausanne, Switzerland
\\
 \small{
   \textbf{Correspondence:} \href{mailto:email@domain}{lucile@ellisalicante.org}
 }}

\begin{document}

\maketitle
\begin{abstract}
The widespread adoption of chat interfaces based on Large Language Models (LLMs) raises concerns about promoting superficial learning and undermining the development of critical thinking skills. Instead of relying on LLMs purely for retrieving factual information, this work explores their potential to foster deeper reasoning by generating critical questions that challenge unsupported or vague claims in debate interventions. This study is part of a shared task of the 12th Workshop on Argument Mining, co-located with ACL 2025, focused on automatic critical question generation. We propose a two-step framework involving two small-scale open source language models: a \texttt{Questioner} that generates multiple candidate questions and a \texttt{Judge} that selects the most relevant ones. %The relatively small size and open nature of the LLMs ensures accessibility, preserves user privacy, and allows for efficient and local deployment, which is particularly valuable in educational settings.
Our system ranked first in the shared task competition, demonstrating the potential of the proposed LLM-based approach to encourage critical engagement with argumentative texts. 
\end{abstract}

\section{Introduction} \label{sec:intro}
\begin{figure*}[ht]
\centering
\includegraphics[width=0.9\textwidth]{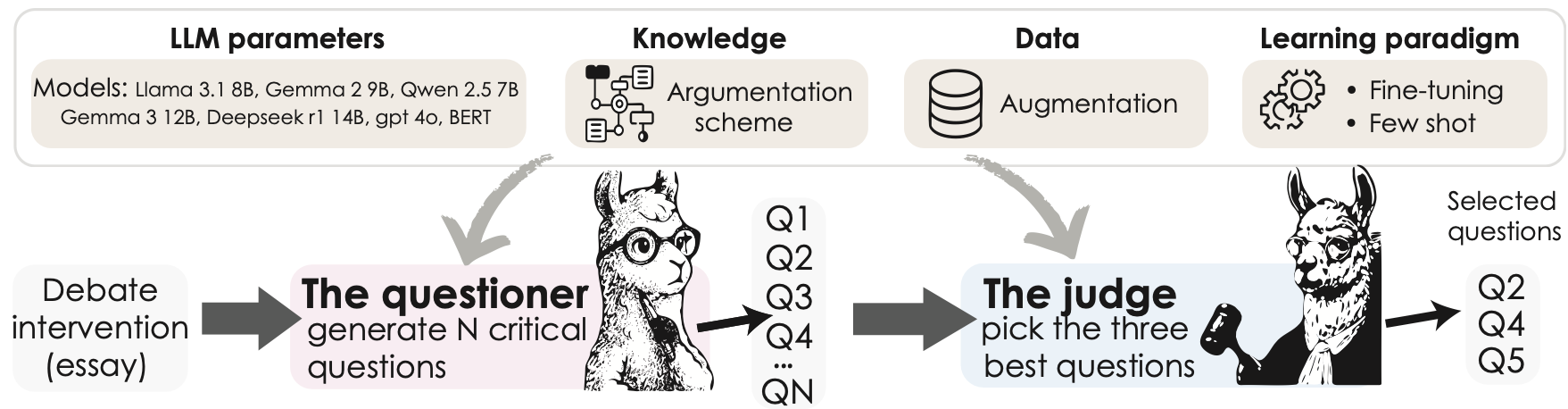}
\caption{Overview of the proposed framework. Given a debate intervention as input, a first LLM, the \texttt{Questioner} $LLM_Q$, generates several candidate questions, and a second LLM, the \texttt{Judge}, $LLM_J$, chooses the three most useful critical questions among the questions generated.}
\label{fig:visual}
\end{figure*}

The intensive use of chatbots based on Large Language Models (LLMs) has been associated with the promotion of superficial learning habits and a decline in critical thinking skills in their users, particularly students \cite{gerlich2025ai,schei2024perceptions}. Motivated by this fact, rather than relying on LLMs to provide factual answers, there is an opportunity to leverage the sophisticated natural language understanding capabilities of LLMs to foster critical thinking by means of the generation of critical questions.

This paper contributes to the CQs-Gen shared task of the 12th Workshop on Argument Mining, co-located with ACL 2025, which focuses on generating critical questions from debate interventions \cite{calvo2025argminingsharedtask}. While previous research has extensively explored the automatic generation of questions \cite{mulla2023automatic, ling2024automatic}, and current AI systems are capable of detecting misinformation with reasonable accuracy \cite{guo2022survey}, relatively little work has leveraged argumentation theory to identify missing claims and misinformation in argumentative text \cite{figueras2024using}, and to generate relevant critical questions about the text \cite{favero2024enhancing, figueras2024critical, ruiz2025explainable}.

To fill this gap, we present an LLM-based framework for generating critical questions from argumentative text, aimed at encouraging users to reflect before accepting a claim. Our approach uses relatively small, open-source\footnote{We use the term ``small'' to refer to LLMs in the 7B-14B range, able to run on a student's laptop, and ``open-source'' to refer to LLMs that are freely available with at least open weights.} LLMs to generate critical questions from a given debate intervention. Figure~\ref{fig:visual} illustrates the pipeline of the proposed method, detailing each step in the process.
 
In sum, the main contributions of our work are four-fold: 
(1) We propose a \textbf{Two-Step Framework for Critical Question Generation} composed of a \texttt{Questioner–Judge} LLM architecture where the \texttt{Questioner}, $LLM_Q$, generates multiple candidate questions that are evaluated by the \texttt{Judge}, $LLM_J$, which selects the most relevant ones, improving quality through selection; (2) we perform an extensive \textbf{empirical evaluation} of several small (7B–14B), open-source LLMs, demonstrating their strong performance despite limited size and without fine-tuning; (3) we explore how integrating 
\textbf{argumentation scheme theory} into prompts —both selectively and systematically— impacts generation quality and diversity; and (4) we highlight the potential of the proposed method to support \textbf{educational tools} that can be deployed locally, preserving privacy and reducing computational costs. 

Our system ranked first in the CQs-Gen 2025 shared task on critical question generation, validating the effectiveness of the proposed approach.
    
\section{Problem definition} \label{sec:litreview}
    
    \subsection{Dataset description}%_____________________________ 
    
The provided dataset \cite{arxivBenchmarkingCritical}, $D_{\text{shared train}}$, is composed of: 

\vspace{0.6ex}
(1) $D = 189$ \textbf{interventions} during real political debate where each intervention consists of a short text of an average of $138 \pm 88.4 $ words; %\\

\vspace{0.6ex}

(2) Their associated \textbf{argumentative schemes}, that is ``stereotypical patterns of inference that capture common types of defeasible arguments, \emph{i.e.}, arguments that are plausible but open to rebuttal. Each scheme represents a form of reasoning with typical premises and a conclusion'' \cite{walton2008argumentation}.\footnote{See \ref{sec:template} for a comprehensive list of the annotation schemes included in the dataset.} Most ($62.4 \%$) of the interventions are associated with a single argumentative scheme, although some may have up to six; %\\ %116 / 186

\vspace{0.6ex}

(3) A set $\mathcal{R}^j$ consisting of $N^j$ annotated \textbf{reference questions} for each debate intervention $j$, where $j = 1 \ldots D$. 
Each reference question $q^j_i$ is labeled with a label or category $l^j_i \in \{\text{Useful}, \text{Unhelpful}, \text{Invalid}\}$, such that $\mathcal{R}^j = \{ (q^j_i, l^j_i) \mid i = 1,\ldots,N^j \}$.
\textbf{Useful} questions can potentially challenge one of the arguments in the text;  \textbf{Unhelpful} questions are valid but unlikely to challenge any of the arguments in the text; and \textbf{Invalid} questions cannot be used to challenge any argument in the intervention \cite{calvo2025argminingsharedtask}.
    
\subsection{Task description}%_____________________________
    The task consists of automatically generating three \textit{Useful} critical questions, $Qc^j = \{qc^j_1, qc^j_2, qc^j_3\}$ for each debate intervention $j$. In this context, critical questions are designed to evaluate the strength of an argument by revealing the assumptions underlying its premises \cite{figueras2024critical}.
    The usefulness of each generated critical question $qc^j_i$ is evaluated by measuring its cosine similarity with the annotated reference questions $\mathcal{R}^j$. The label assigned to $qc^j_i$ corresponds to the label of the most similar reference question, provided that it is larger than or equal to $0.6$. If no similarity score exceeds this threshold, the question is marked as \textit{Not able to evaluate}. In this case, human evaluators assessed the usefulness of the question during the competition. The final score was computed on the $34$ interventions that composed the test set, $D_{\text{shared test}}$. Note that the reference test set, with the labels corresponding to the interventions in $D_{\text{shared test}}$ was not made available.

\section{Methodology}\label{sec:method}

    As illustrated in Figure~\ref{fig:visual}, the proposed system consists of two large language models (LLMs) used sequentially. (1) The \texttt{Questioner} ($LLM_Q$) which generates candidate critical questions given an intervention and its associated argumentation schemes; and the (2) The \texttt{Judge} ($LLM_J$) which evaluates these candidates and selects those deemed most useful \cite{li2024generation}.
    This architecture is grounded in the framework of critical thinking proposed by \citet{elder2020critical}, which comprises analytic, creative, and evaluative dimensions. We operationalize the creative components through $LLM_Q$ (generation), and the analytic and evaluative components through $LLM_J$ (selection).

    \subsection{The prompts}  
    The prompts provided to the LLMs %$LLM_Q$ (resp. $LLM_J$) 
    include: the intervention text, the role of the LLM (\emph{i.e.}, \texttt{Questioner} or \texttt{Judge}), definitions of critical question and argumentation scheme, the argumentative schemes present in the intervention along with their definitions (see ~\ref{sec:template}) and corresponding question templates (see ~\ref{sec:cq_template}), the task objective, and the expected output. For more details, see \ref{sec:prompt}.

    For $LLM_Q$, each prompt is designed to elicit $N$ questions in a single generation step, rather than prompting the model $N$ times for one question at a time. This strategy effectively reduces question repetition.

    Aligned with \citet{guo2023diversifying}, we hypothesize that candidate questions exhibiting high similarity are likely to be useful. Thus, the following instruction is added to $LLM_J$'s prompt: \textit{If some questions are redundant, these questions must be important: select the most relevant one.} This modification led to an overall improvement in performance.
        
    \subsection{Experimental design}
    We split $D_{\text{shared train}}$ into training ($D_{\text{train}}$, 74), validation ($D_{\text{val}}$, 33), and test ($D_{\text{test}}$, 79) sets. The size of $D_{\text{test}}$ was selected to ensure stable results under the automatic evaluation metric (see \ref{sec:limi_eval}).
    We conducted experiments on $D_{\text{test}}$ by varying the following parameters to assess their impact on performance: the choice of LLM for each of the roles (\texttt{Questioner} and \texttt{Judge}), the number of  candidate questions generated, and the temperature setting of the LLMs. Additionally, we performed an ablation study to evaluate the role of argumentation schemes in the generation process and address the added value of $LLM_J$ by comparing it with alternative question selection strategies. For more details on the experimental setup and further experiments, including LLM and BERT fine-tuning and data augmentation, see \ref{sec:exp_setup} and \ref{sec:further}.
        
\section{Experiments and results}\label{sec:res}
    
    \subsection{Model comparison}
    We evaluated both $LLM_Q$ and $LLM_J$ using a selection of small, open-source LLMs ranging from 7B to 14B parameters: Qwen 2.5 7B \cite{yang2024qwen2}, Llama 3.1 8B \cite{dubey2024llama}, Gemma 2 9B \cite{team2024gemma}, Gemma 3 12B \cite{team2025gemma}, and DeepSeek R1 14B \cite{guo2025deepseek} \footnote{For further details, see Section~\ref{sec:other_models}.}. We compare their performance with that of GPT-4o \cite{achiam2023gpt}. 
    
    As shown in Table~\ref {tab:res_model}, the LLM combination yielding the highest proportion of useful outputs on $D_{\text{test}}$ is Llama 3.1 8B as $LLM_Q$ and Gemma 2 9B as $LLM_J$.

    \begin{table}[ht]
    \centering
    {\small
    \resizebox{\columnwidth}{!}{%
    \begin{tabular}{|P{1.6cm}P{1.6cm}|p{0.9cm}p{0.9cm}p{0.9cm}|}
    \hline
    \boldmath{\textbf{$LLM_Q$}} & \boldmath{\textbf{$LLM_J$}} & \textbf{\small{Use.}} $\uparrow$  & \textbf{\small{Inv.}} $\downarrow$ & \textbf{\small{NoEval} } \\
    \hline
    Llama 3.1   & -               & $53.2$ & $3.0$ & $33.8$ \\
    Gemma 2   & -               &$46.4$ &$3.0$  & $42.6$  \\
    Gemma 3   & -               & $40.5$ & $\boldmath{\textbf{2.5}}$  & $46.0$ \\
    Llama 3.1  & Llama 3.1       &  $53.2$& $3.9$ & $33.0$  \\
    Llama 3.1  & Qwen 2.5      & $56.8$ & $3.3$  & $28.6$ \\
    Llama 3.1  & Gemma 2      & \boldmath{\textbf{$57.6$}} & $5.2$ &$30.3$  \\
    Llama 3.1  & Gemma 3      & $57.1$ & $2.6$ & $30.7$ \\
    \hline 
    \end{tabular}}}
\caption{\textbf{Performance on $D_{\text{test}}$ for a selection of $LLM_Q$ and $LLM_J$.} \textit{Use}, \textit{Inv} and \textit{NoEval} are the $\%$ of \textit{Useful}, \textit{Invalid}, and \textit{Not able to evaluate} questions, respectively. $LLM_Q$ generates $8$ questions of which $LLM_J$ selects the best $3$. The argumentative schemes are not given in the prompt. Best results in bold.}
    \label{tab:res_model}
    \end{table}
    %1 20250325225900_llama3.1__R1_sq79_G_EG0_Q8
    %3 20250326021404_gemma2__R1_sq79_G_EG0_Q8
    %4 20250326100422_gemma3_12b__R1_sq79_G_EG0_Q8
    %6 20250325231346_llama3.1llama3.1_R1_sq79_G_C_EG0_Q8
    %7  20250320235141_llama3.1qwen2.5_R2_sq79_G_C_EG0_Q8
    %8 20250325233631_llama3.1gemma2_R1_sq79_G_C_EG0_Q8
    %9 20250326070038_llama3.1gemma3_12b_R1_sq79_G_C_EG0_Q8
    %10 20250321012241_llama3.1deepseek-r1_14b_R2_sq79_G_C_EG0_Q8  
\vspace*{-.15in}
    \subsection{Leveraging argumentation schemes}
   To assess the impact on performance of adding argumentation scheme theory in the prompts for both $LLM_Q$ and $LLM_J$, we conducted an ablation study. Table~\ref{tab:res_sch} compares the performance of $LLM_Q$ (Llama 3.1 8B, generating six questions) and $LLM_J$ (Gemma 2 9B) with the following configurations: (\textbf{1) Without}: No argumentation scheme is provided; \textbf{(2) With (one)}: All argumentation schemes relevant to the given intervention are included in a single prompt; \textbf{(3) With (mult.)}: Each argumentation scheme is provided in a separate prompt; and \textbf{(4) Both }: $LLM_Q$ is prompted independently using the \textbf{With (one) } and without argumentation schemes setups. Then the two sets of candidate questions are merged for their selection by $LLM_J$ . Similarly to previous work \citep{figueras2024critical}, the best performance is achieved in the \textbf{Both} configuration, suggesting that combining scheme-based and non-scheme-based prompts yields the most effective results. Note that $81\%$ of the questions selected by $LLM_J$ were generated with the argumentation scheme in the prompt. 
        \begin{table}[ht]
        \centering
        \renewcommand{\arraystretch}{0.8}
        \resizebox{\columnwidth}{!}{%
        \begin{tabular}{|>{\centering\arraybackslash}m{2cm}|p{0.8cm}p{0.8cm}p{0.8cm}|p{0.8cm}p{0.8cm}p{0.8cm}|}
            \hline
            \multirow{2}{*}{\textbf{Scheme}} & \multicolumn{3}{c|}{\textbf{$LLM_Q$}} & \multicolumn{3}{c|}{\textbf{$LLM_Q + LLM_J$}} \\
            \cline{2-7}
            &\small{Use.}$\uparrow$  & \small{Inv.} $\downarrow$ & \small{NoEval} & \small{Use.}$\uparrow$  & \small{Inv.} $\downarrow$ & \small{NoEval} \\
            \hline
            \textbf{Without} &$54.7$ &$3.2$ & $32.7$& $57.7$&$3.8$  & $27.4$  \\
            \textbf{With (one)}  & $53.4$ & $4.0$&$32.1$ & $56.5$ &$3.7$  &$28.3$ \\
            \textbf{With (mult.)} & $46.0$& $4.0$&$27.0$ &$51.6$ &$3.6$  & $34.2$\\
            \textbf{Both} & $54.0$ &$3.5$ & $31.0$&\boldmath{\textbf{$62.4$}} &\boldmath{\textbf{$2.1$}}  &\boldmath{\textbf{25.7}}  \\
            \hline
        \end{tabular}%
        }
        \caption{\textbf{Performance on $D_{\text{test}}$ with different argumentation schemes setups.} $LLM_Q$: Llama 3.1 generating $6$ questions. \textit{Without}: No argumentation scheme is provided;\textit{With (one)}: Argumentation schemes are included in a single prompt; \textit{ With (mult.)}: Each argumentation scheme is provided in a separate prompt; and \textit{Both}: $LLM_Q$ is prompted independently with and without argumentation schemes. Best results in bold.}
        \label{tab:res_sch}
        \end{table}
        % 1 20250328224153_llama3.1gemma2_R1_sq79_G_C_EG0_Q6no_sch
        % 2 20250331000941_llama3.1gemma2_R1_sq79_E_C_EG0_Q6redund
        % 3 20250401234308_llama3.1gemma2_R1_sq76_E_C_EG0_Q6one_sch
        % 4 20250330233232_llama3.1gemma2_R1_sq79_G_E_C_EG0_Q6redund
        % mean of sch: 0.5506  0.337 0.289 and llm: 0.5295 0.0359 0.3249
    
    \subsection{Number of candidate questions}
    Table~\ref{tab:res_quest} presents the effectiveness of the questions as a function of the number of candidate questions generated per prompt. The experiment uses Llama 3.1 8B as $LLM_Q$—prompted both with and without the schemes—and Gemma 2 9B as $LLM_J$. Generating four candidate questions per prompt (eight in total) yielded the best performance.
    \begin{table}[ht]
    \centering
    \resizebox{\columnwidth}{!}{%
    \begin{tabular}{|>{\centering\arraybackslash}m{1.4cm}|p{2.3cm}p{2.3cm}p{2.3cm}|}
        \hline
        \textbf{\# quest.} & \makebox[2.3cm][c]{\textbf{\small Use.\,$\uparrow$}}  
        & \makebox[2.3cm][c]{\textbf{\small Inv.\,$\downarrow$}} 
        & \makebox[2.3cm][c]{\textbf{\small NoEval}} \\
        \hline
        \textbf{4} & \makebox[2.3cm][c]{\boldmath{$59.3 \pm 3.36$}} 
        & \makebox[2.3cm][c]{\boldmath{$2.80 \pm 1.05\mathrm{e}^{-1}$}} 
        & \makebox[2.3cm][c]{\boldmath{$22.5 \pm 2.13$}} \\
        \textbf{6} & \makebox[2.3cm][c]{$57.2 \pm 8.82\mathrm{e}^{-1}$}  
        & \makebox[2.3cm][c]{$2.72 \pm 6.38\mathrm{e}^{-1}$}
        & \makebox[2.3cm][c]{$25.9 \pm 2.91$} \\
        \textbf{8} & \makebox[2.3cm][c]{$57.3 \pm 7.58\mathrm{e}^{-1}$}  
        & \makebox[2.3cm][c]{$3.22 \pm 2.78\mathrm{e}^{-1}$} 
        & \makebox[2.3cm][c]{$25.7 \pm 8.99\mathrm{e}^{-1}$} \\
        \hline
    \end{tabular}%
    }
    \caption{\textbf{Performance on $D_{\text{Shared\_train}}$ as a function of the number of candidate questions generated.} $LLM_Q$: Llama 3.1, $LLM_J$: Gemma 2. 3 runs.}
    \label{tab:res_quest}
    \end{table}
    % 1 20250404175730_llama3.1gemma2_R3_sq186_G_E_C_EG0_Q4test_Q
    % 2 20250404215804_llama3.1gemma2_R3_sq186_G_E_C_EG0_Q6test_Q
    % 3 20250405020639_llama3.1gemma2_R3_sq186_G_E_C_EG0_Q8test_Q
    \vspace*{-.2in}
    \subsection{Added value of the \texttt{Judge}, $LLM_J$}
    % 1 2 20250404175730_llama3.1gemma2_R3_sq186_G_E_C_EG0_Q4test_Q
    Although we observe an improvement in performance when adding $LLM_J$ versus a random selection (see Tables~\ref{tab:res_sch} and \ref{tab:res_quest}), the results are not directly comparable, as the average usefulness is computed over different numbers of questions ($N$ for $LLM_Q$ and three for $LLM_J$). To further assess the effectiveness of $LLM_J$, we compared it against alternative selection paradigms.
    Table~\ref{tab:res_judge} reports the performance $LLM_J$ versus a selection by an oracle and randomly, using Llama 3.1 8B as $LLM_Q$ with four candidate questions per prompt. The oracle selects up to three useful questions. If fewer than three \textit{Useful} questions are available, the remaining slots are filled by \textit{Unhelpful} questions. If still insufficient, \textit{Invalid}, and then \textit{Not able to evaluate} questions are considered, in that order. The oracle illustrates the upper bound of the \texttt{Judge}’s potential performance.
    Results show that $LLM_J$ achieves a usefulness rate that is $3.4$ percentage points higher than random selection, a statistically significant improvement ($p<0.05$, McNemar's test). As expected, the oracle yields the highest usefulness with a gain of $34.2$ percentage points.
  \begin{table}[ht]
    \centering
    \renewcommand{\arraystretch}{0.99}
    \resizebox{\columnwidth}{!}{%
    \begin{tabular}{|>{\centering\arraybackslash}m{1.9cm}|p{2.3cm}p{2.3cm}p{2.3cm}|}
    \hline \textbf{Selection} &\makebox[2.2cm][c]{\textbf{\small{Use.} $\uparrow$}}  & \makebox[2.2cm][c]{\textbf{\small{Inv.}} $\downarrow$} & \makebox[2.2cm][c]{\textbf{\small{NoEval}}} \\
    \hline
    \textbf{Random}   & $55.9 \pm 2.22$ & $2.7 \pm 7.94 \mathrm{e}^{-3}$& $25.7 \pm 2.5\mathrm{e}^{-2}$  \\
     \textbf{Gemma 2}   &$59.3 \pm 3.36$ & $2.8 \pm 1.05\mathrm{e}^{-1}$ & $22.5 \pm 2.13$\\
    \textbf{Oracle}  & $93.5 \pm 1.19$ & $6.68 \pm 8.59\mathrm{e}^{-2}$ & $1.70 \pm 8.19\mathrm{e}^{-1}$ \\
    \hline
    \end{tabular}}
    
    \caption{\textbf{Performance on $D_{\text{Shared train}}$ depending on the method to select the questions.} Comparison between random selection, selection with Gemma 2 as $LLM_J$ or with an Oracle. In all cases, $LLM_Q$ is Llama 3.1 generating $4+4$ questions. 3 runs.}
    \label{tab:res_judge}
    \end{table}
    
    \subsection{Final submission}   
    
     Based on the results of the previous experiments, we selected the following setup for our final submission: $LLM_Q$, Llama 3.1 8B, generating four questions without the scheme and four with the scheme, all within a single prompt; $LLM_J$, Gemma 2 9B, selecting the three best questions, used without fine-tuning. 
    For comparison, we maintained the same experimental setup but substituted $LLM_J$ with GPT-4o in our second submission, and in the third submission, GPT-4o was used for both $LLM_Q$ and $LLM_J$ under identical prompting conditions.  Table~\ref {tab:res_final} shows the performance with the automated evaluation on $D_{\text{Shared train}}$ and $D_{\text{Shared test}}$ for the three final submissions.
    
\begin{table}[ht]
    \centering
    {\footnotesize 
    \renewcommand{\arraystretch}{1.2}
    \setlength{\tabcolsep}{4pt}

    \begin{tabular}{|>{\centering\arraybackslash}m{0.8cm}|p{1cm}p{1cm}|p{1cm}p{1cm}|}
        \hline
        \multirow{2}{*}{\textbf{Sub.}} & \multicolumn{2}{c|}{\textbf{Valiadation}} & \multicolumn{2}{c|}{\textbf{Test}} \\
        \cline{2-5}
         & \textbf{Use.\,$\uparrow$} & \textbf{NoEval} & \textbf{Use.\,$\uparrow$} & \textbf{NoEval} \\
        \hline
        1 & 61.4 & 21\% & 36.3 & 36\% \\
        2 & 61.0 & 19\% & 44.1 & 36\% \\
        3 & \textbf{64.4} &\textbf{ 19\%} &\textbf{ 50.0} & \textbf{28\%} \\
        \hline
    \end{tabular}
    }
    \caption{\textbf{Performance with the automated evaluation on the validation set and the test set for the three final submissions.} Bold indicates the winning submission.}
    \label{tab:res_final}
\end{table}

After the manual annotation of the questions by the organizers, the score of the best performing submission rose to \textbf{67.6}, ranking first in the task.
    
\section{Discussion and conclusion}\label{sec:discussion}
In this paper, we have proposed a two-step framework for generating critical questions, where one LLM ($LLM_Q$) generates multiple candidate questions and another LLM ($LLM_J$) evaluates and selects the most relevant ones \cite{li2024generation}. This selection-based approach consistently outperformed direct generation, emphasizing the benefits of separating generation from evaluation. 
Our experiments show that adding argumentation schemes to the prompts improves the quality of the generated questions. However, strictly enforcing these schemes can reduce diversity. Thus, a selective use of schemes strikes a better balance between structural guidance and creative generation, in line with prior work \cite{figueras2024critical}.

Given the small size of the dataset, traditional strategies such as fine-tuning or data augmentation (\emph{e.g.}, using BERT-based methods) yielded limited improvement. Instead, leveraging small, open-source LLMs 
guided by domain-specific argumentation theory proved more effective in this low-resource setting.

\section*{Limitations} \label{sec:lim}
This work has several limitations that should be acknowledged. 

The first limitation concerns the evaluation methodology, which relies on an automatic comparison with a set of predefined reference questions by means of cosine similarity. Many generated questions did not align with any reference, despite being potentially useful, and hence were labeled as \textit{Not able to evaluate}. This mismatch introduces a risk of mis-estimation of the model's performance and could lead to overfitting.

A second limitation stems from the use of a small, domain-specific dataset focused on political discourse, which at times lacks sufficient context for effective question generation. This narrow scope limits the generalizability of our findings. Future work should aim to evaluate the proposed framework on broader and more diverse datasets to assess its robustness across different domains, like education.

A third limitation lies in the performance of $LLM_J$, which shows a substantial gap compared to the oracle. This indicates that while some generated questions are \textit{Useful}, the \texttt{Judge} does not consistently identify them, suggesting significant potential for future improvements.

\section*{Acknowledgements}
L.F. and N.O. have been partially funded by a nominal grant received at the ELLIS Unit Alicante Foundation from the Regional Government of Valencia in Spain (Convenio Singular signed with Generalitat Valenciana, Conselleria de Innovación, Industria, Comercio y Turismo, Dirección General de Innovación). L.F. has also been partially funded by a grant from the Banc Sabadell Foundation.

%\bibliography{latex/custom}

\appendix
 
\section{Appendix}
    \subsection{The datasets}
\begin{figure}[h!]
        \centering
        \includegraphics[width=0.46\textwidth]{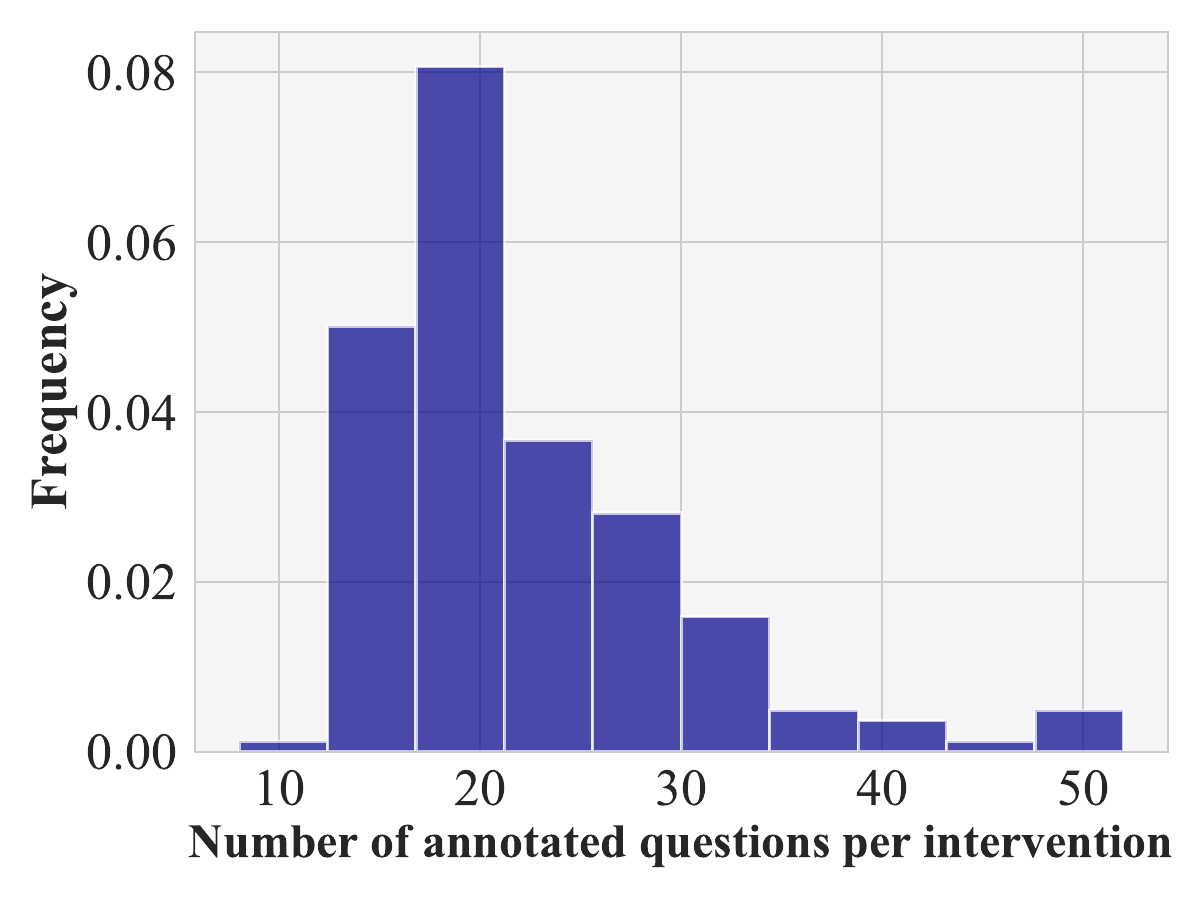}
        \caption{Distribution of the number of annotated questions per intervention in $D_{\text{shared train}}$.}
        \label{fig:anno}
        \end{figure}

\begin{figure}[h!]
        \centering
        \includegraphics[width=0.46\textwidth]{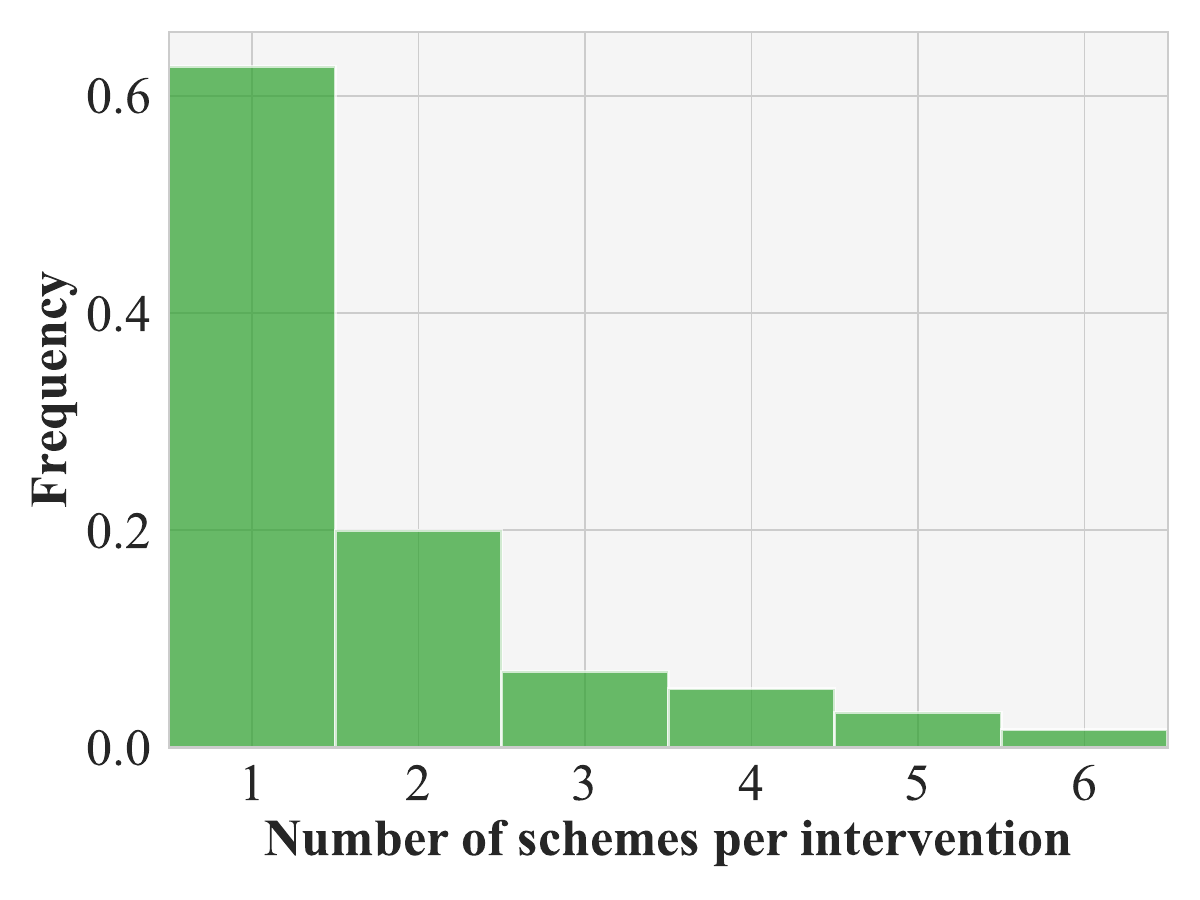}
        \caption{Distribution of the number of schemes per intervention in $D_{\text{shared train}}$.}
        \label{fig:sch}
        \end{figure}

\begin{figure}[h!]
        \centering
        \includegraphics[width=0.46\textwidth]{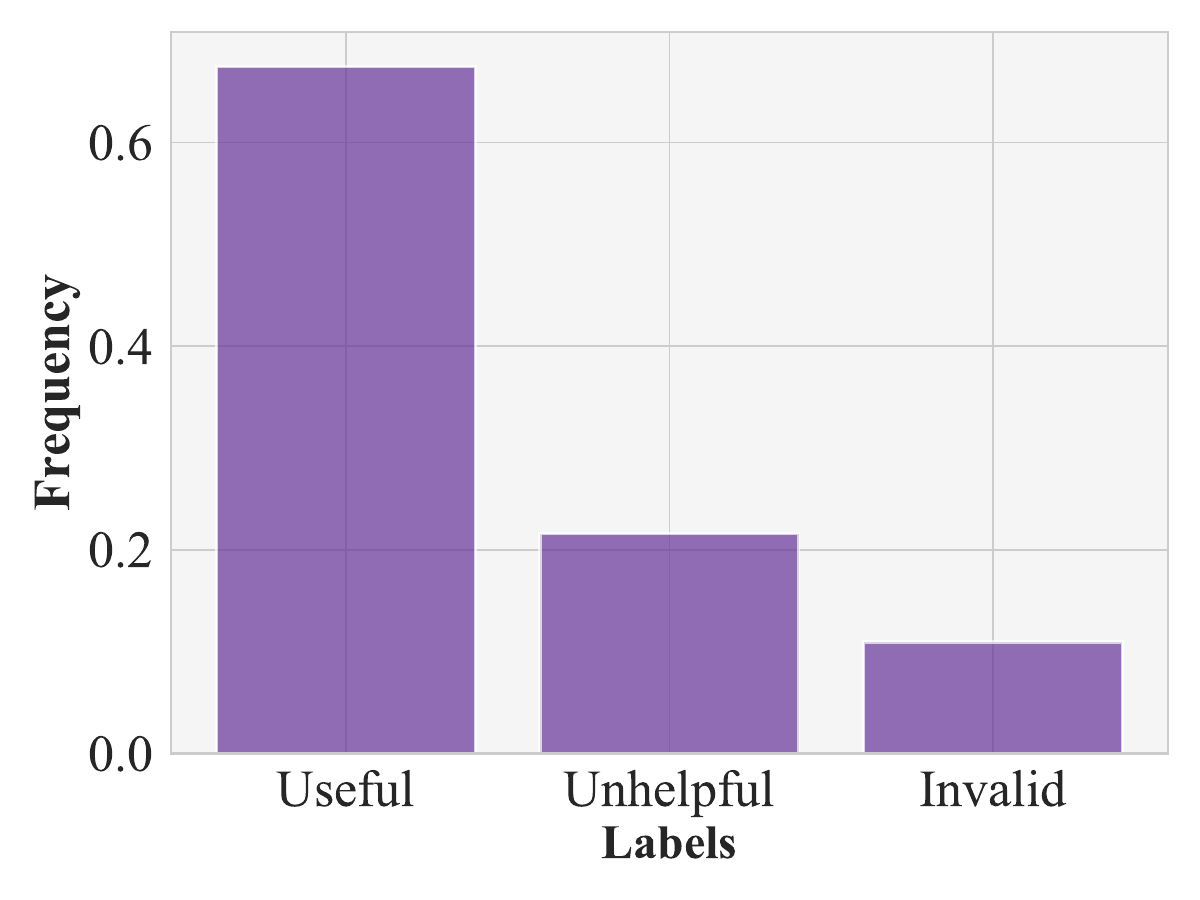}
        \caption{Distribution of the labels in $D_{\text{shared train}}$.}
        \label{fig:label}
        \end{figure}

        Figure~\ref{fig:anno} shows the distribution of the number of annotated questions per intervention in $D_{\text{shared train}}$, Figure~\ref{fig:sch} shows the distribution of the number of schemes per intervention in $D_{\text{shared train}}$ and Figure~\ref{fig:label} shows the distribution of the labels in $D_{\text{shared train}}$.
        
    \subsection{Experimental setup}\label{sec:exp_setup}
        \paragraph{Data split}
        To enable training (see \ref{sec:further}), we divided $D_{\text{shared train}}$ (189 interventions) into three subsets: a training set $D_{\text{train}}$ (74 interventions), a validation set $D_{\text{val}}$ (33 interventions), and a test set $D_{\text{test}}$ (79 interventions).  The test set was intentionally large to accommodate the variability due to the automatic evaluation metric (see \ref{sec:limi_eval}), aiming for more stable and reliable results.
        
        \paragraph{Software and hardware setup}
        All experiments were performed on an Apple M1 Pro laptop with 32 GB RAM using Ollama \footnote{\url{https://github.com/ollama/ollama},\\ \url{https://ollama.com.}}, an open-source framework that enables users to run, create, and share LLMs locally on their machines.
        Our code is available at \url{https://github.com/lucilefavero/SQ_shared_task}.

        \subsubsection{Limitation of the automatic evaluation and human evaluation} \label{sec:limi_eval}
         Due to the nature of the automatic evaluation metric, a substantial proportion of generated questions could not be evaluated (see the column \textit{No} in Tables~\ref{tab:res_model}, \ref{tab:res_sch}, and \ref{tab:res_quest}). Consequently, distinguishing the best-performing configurations was not straightforward. The performance differences across varying numbers of candidate questions were small and overshadowed by significant variance in our test set $D_{\text{test}}$. To improve precision, we repeated the experiment three times on $D_{\text{shared train}}$ (the entire dataset).

        Throughout our experiments, we prioritized the quality of critical questions over minimizing the proportion of \textit{Not able to evaluate} labels. We intentionally avoided overfitting to the automatic evaluation metric, under the assumption that some unevaluated questions might still be useful. The primary goal was to reduce the proportion of \textit{Invalid} and \textit{Unhelpful} labels. To ensure quality, we manually evaluate some questions that could not be evaluated by the automatic scoring system.
        
        \subsubsection{Additional information on the LLMs}\label{sec:other_models}
        A complete list of the tested models is provided below. The results for the most relevant combinations of $LLM_Q$ and $LLM_J$ are presented in Table~\ref{tab:res_model}.
        \begin{itemize} 
        \item  \textbf{Qwen 2.5 7B}. Qwen 2.5 is a multilingual transformer-based LLM with RoPE, SwiGLU, RMSNorm, and Attention QKV bias, released in September 2024 by the Qwen Team. \cite{yang2024qwen2}. 
        \item \textbf{Llama 3.1 8B}, Llama 3.1 is a multilingual large language model optimized for dialogue applications. It supports eight languages and offers a context window of up to 128,000 tokens, enabling it to handle extensive conversational contexts. Released in July 2024 by Meta \cite{dubey2024llama}. 
        \item  \textbf{Gemma 2 9B}, Gemma 2 is a text-to-text decoder-only LLM available in English with open weights, released in June 2024 by Google, \cite{team2024gemma}.
        \item  \textbf{Gemma 3 12B}, Gemma 3 is another model from the Gemma family; it has longer context, a different architecture than Gemma 2, and is trained with distillation. It was released in March 2025 by Google, \cite{team2025gemma}.
        \item \textbf{DeepSeek R1 14B}. DeepSeek R1 is an open-source large language model designed to enhance reasoning capabilities through reinforcement learning. It rivals other advanced models in tasks such as mathematics, coding, and logical reasoning. Released in January, 2025 by the Chinese AI startup DeepSeek \cite{guo2025deepseek}.
        \end{itemize}
    
        \subsubsection{Structure of LLM's prompts} \label{sec:prompt}
        
        The prompt for $LLM_Q$ consists of the following components:
        \begin{itemize}
        \item \textbf{The intervention}. 
        \item \textbf{Role.} ``You are a critical judge.''
        \item \textbf{Definition of critical question}: ``Critical questions are the set of enquiries that should be asked in order to judge if an argument is good or fallacious by unmasking the assumptions held by the premises of the argument.''
        \item \textbf{Definition of argumentation scheme.} ``Argumentative schemes are stereotypical patterns of inference that capture common types of defeasible arguments, i.e. arguments that are plausible but open to rebuttal. Each scheme represents a form of reasoning with typical premises and a conclusion.''
        \item \textbf{The argumentation schemes present in the intervention with their definition and template of critical questions} see \ref{sec:template}
                \item \textbf{Goal.} ``Use the provided scheme and template of critical questions to generate [N] critical questions to evaluate the arguments in the given essay.''
        \item \textbf{Expected output.} ``Give one question per line. Make the questions simple, and do not give any explanation regarding why the question is relevant.''

        \end{itemize}
        
        The prompt for $LLM_J$ consists of the following components:
        \begin{itemize}
        \item \textbf{The intervention}. 
        \item \textbf{Role.} ``You are a very strict critical and sceptical judge.''
        \item \textbf{Definition of critical question}: ``Critical questions are the set of enquiries that should be asked in order to judge if an argument is good or fallacious by unmasking the assumptions held by the premises of the argument.''
        \item \textbf{Definition of argumentative scheme.} ``Argumentative schemes are stereotypical patterns of inference that capture common types of defeasible arguments, i.e. arguments that are plausible but open to rebuttal. Each scheme represents a form of reasoning with typical premises and a conclusion.''
        \item \textbf{The argumentation schemes present in the intervention with their definition and template of critical questions} see \ref{sec:template}
                \item \textbf{Goal.} ``Select the 3 best critical questions that should be raised before accepting the arguments in the essay. If some questions are redundant, these questions must be important: select the most relevant one.''
        \item \textbf{Expected output.} ``Give one question per line. Make the questions simple, and do not give any explanation regarding why the question is relevant.''

        \end{itemize}

    \subsection{Argumentation scheme definition and template}\label{sec:template}
    Table~\ref{tab:template} depicts the argumentation schemes identified in the dataset, along with their corresponding critical question templates. The definitions and templates are adapted from \citet{walton2008argumentation}.

    \subsection{ Further experiments} \label{sec:further}
        \paragraph{Critical questions' templates} \label{sec:cq_template}
        We examined two approaches for incorporating critical question templates into the prompts: using the template provided by \citet{figueras2024critical}, and utilizing the critical question templates outlined in Table~\ref{tab:template}. We noted a slight performance improvement with the first template for the configurations employing $LLM_Q$: Llama 3.1, $LLM_J$: Gemma 2 or GPT-4o, and with the second template for the configurations involving $LLM_Q$: GPT-4o and $LLM_J$: GPT-4o. This approach was adopted in our final submission.
        
        \paragraph{Temperature}
        We also explored modifying the generation temperature from its default setting and observed an overall decrease in performance.
        
        \paragraph{Fine-tuning}
        Attempts to fine-tune both $LLM_Q$ and $LLM_J$ were inconclusive. The resulting model outputs often diverged significantly from the intended instructions and demonstrated poor performance, likely attributable to task complexity combined with the limited size of the training dataset  $D_{\text{Train}}$.

        We also fine-tuned BERT \cite{devlin2018bert} to classify candidate questions into three categories: \textit{Useful}, \textit{Unhelpful}, and \textit{Invalid}, selecting the three questions with the highest predicted probability of being \textit{Useful}. However, similar to the LLM fine-tuning, the model failed to outperform a random baseline, likely due to task complexity and the limited size of $D_{\text{Train}}$.
        
        \paragraph{Data augmentation}
        The poor performance of the trained approaches is likely attributable to the limited and highly imbalanced annotated dataset. Specifically, over $67\%$ of annotations in $D_{\text{Shared train}}$ are labeled as \textit{Useful} (see Figure~\ref{fig:label}). To address this, we augmented $D_{\text{Train}}$ with Llama 3.1, generating questions and matching them to reference annotations to balance label distribution across interventions. However, fine-tuning the LLMs and BERT  on the augmented data still yielded inconclusive results.

   \subsection{Further comments on the results } 
    The automatic evaluation scores on $D_{\text{shared test}}$ are lower compared to those obtained on $D_{\text{shared train}}$, likely due to increased variance arising from the test set's smaller size. Additionally, the shared-task organizers indicated, after human evaluation, that the test set presented a higher difficulty.
    \onecolumn
    
    \begin{longtable}{|>{\raggedright}p{2.1cm}|p{5.2cm}|p{7cm}|}
    \hline
    \multicolumn{3}{r}{\textit{Continued from previous page}} \\
    \hline
    \endhead
    
    \hline
    \multicolumn{3}{r}{\textit{Continued on next page}} \\
    \endfoot
    
    \endlastfoot
    \hline
    \textbf{Name} & \textbf{Scheme's definition} & \textbf{Critical questions template} \\ \hline
    \endhead
    Ad Hominem & This scheme attacks an opponent’s argument by alleging inconsistency between their actions and their stated position. & Is the alleged inconsistency real and relevant to the argument? Does the inconsistency undermine the argument’s validity? Could the argument still hold despite the personal inconsistency? \\ \hline
    
    Alternatives & This scheme reasons that one option should be chosen (or avoided) by comparing it to other possible options. & Have all relevant alternatives been considered? Are the alternatives fairly evaluated? Is the chosen alternative clearly superior based on the criteria? \\ \hline
    
    Analogy & This scheme draws a conclusion about one case by comparing it to a similar case where the conclusion is known to hold. & Are the two cases sufficiently similar in relevant respects? Are there significant differences that undermine the analogy? Is the conclusion in the known case well-established? \\ \hline

    Bias & This scheme attacks an argument by alleging that the source is biased, thus undermining its credibility. & Is there clear evidence of bias in the source? Does the alleged bias directly affect the truth of the argument’s conclusion? Could the argument still hold despite the bias? \\ \hline

    Cause to effect & This scheme reasons that if a certain cause occurs, it will lead to a specific effect, based on a causal relationship. & Is there sufficient evidence that the cause reliably produces the effect? Could other factors intervene to prevent the effect from occurring? Is the causal link based on correlation rather than proven causation? \\ \hline
    
    Consequences & This scheme bases a conclusion on the positive or negative outcomes of a proposed action, arguing for or against it based on those consequences. & Are the predicted consequences likely to occur if the action is taken? Are there other consequences (positive or negative) that haven’t been considered? Is the evaluation of the consequences as good or bad justified? \\ \hline

    Example & This scheme involves reasoning from a specific case or instance to a general conclusion, suggesting that what holds in the example applies more broadly. & Is the example representative of the broader category or situation? Are there significant counterexamples that undermine the generalization? Is the example relevant to the conclusion being drawn? \\ \hline

    Expert opinion & This scheme concludes that a proposition is true because an expert in the relevant field asserts it. & How credible is the expert as a source? Is the expert an authority in the field relevant to the proposition? What exactly did the expert assert? Is the expert personally reliable and trustworthy? Is the expert’s claim consistent with other experts? Is the expert’s assertion backed by evidence? \\ \hline

    Fear and danger appeals & This scheme urges action or avoidance based on the fear of a harmful outcome if the action isn’t taken or is taken. & Is the feared outcome realistically likely to occur? Is the fear disproportionate to the evidence of danger? Are there other ways to mitigate the feared outcome without the proposed action? \\ \hline
    
    Negative consequences & This scheme argues against an action because it will lead to bad outcomes. & Are the negative consequences probable? Are there positive consequences that might offset the negative ones? Is the judgment of the consequences as negative reasonable? \\ \hline

    Popular opinion & This scheme argues that a proposition is true or should be accepted because it is widely believed by the majority. & Is the opinion truly held by a significant majority? Does the majority have reliable evidence or expertise to justify their belief? Could the majority be mistaken or influenced by bias? \\ \hline
    
    Popular practice & This scheme justifies an action or belief because it is commonly practiced by many people. & Is the practice widespread enough to be considered popular? Does the practice’s popularity indicate its correctness or value? Are there reasons the practice might be flawed despite its popularity? \\ \hline
    
    Positive consequences & This scheme argues for an action because it will plausibly lead to good outcomes. & Are the positive consequences likely to occur? Are there potential negative consequences that outweigh the positive ones? Is the assessment of the consequences as positive well-founded? \\ \hline

    Position to know & This scheme concludes a proposition is true because the source is in a position to know about it (e.g., firsthand experience). & Is the source genuinely in a position to know about the proposition? Is the source honest and trustworthy? Did the source actually assert the proposition? \\ \hline
    
    Practical reasoning & This scheme involves an agent reasoning from a goal to an action that is a means to achieve that goal (e.g., ``I want G, doing A achieves G, so I should do A''). & What other goals might conflict with G? Are there alternative actions to A that could also achieve G? Is A the most efficient means to achieve G? Is it practically possible for me to carry out A? What are the potential side effects or consequences of doing A? \\ \hline
    
    Sign & This scheme infers a conclusion based on an observable sign or indicator that suggests the presence of a condition or event. & Is the sign a reliable indicator of the conclusion? Could the sign be present without the conclusion being true? Are there alternative explanations for the sign? \\ \hline
    
    Value & This scheme reasons that an action should be taken or avoided because it aligns with or conflicts with an agent’s values (e.g., ``V is good, so I should pursue G that promotes V''). & Is value V genuinely positive/negative as judged by the agent? Does pursuing V conflict with other values the agent holds? Is the link between the action and the promotion of V well-supported? \\ \hline
    
    Verbal classification & This scheme applies a general rule or property to a specific case based on how the case is classified linguistically. & Is the classification of the case accurate and appropriate? Does the general rule reliably apply to all cases under this classification? Is the classification ambiguous or contested? \\ \hline

    \caption{Argumentation schemes identified in the dataset, along with their corresponding critical question templates. Definitions and templates are adapted from \citet{walton2008argumentation}.}
    \label{tab:template}
    \end{longtable}
    \twocolumn
    
\end{document}